\documentclass{article} 
\usepackage{iclr2017_conference,times}
\usepackage{hyperref}
\usepackage{url}

\usepackage{times}
\usepackage{latexsym}
\usepackage{graphicx}
\usepackage[draft]{pgf}
\usepackage{amsmath}
\usepackage{amsfonts}
\usepackage{booktabs}
\usepackage{nicefrac}       
\usepackage{hyperref}       
\usepackage{url}            
\usepackage{enumitem}
\newif\ifshowcomments
\showcommentstrue
\ifshowcomments
\newcommand\sidaw[1]{\textcolor{blue}{[sidaw: #1]}}

\newcommand\TODO[1]{\textcolor{red}{[TODO: #1]}}
\else
\newcommand\sidaw[1]{}
\newcommand\TODO[1]{}
\fi

\newcommand\blanktok{\ensuremath{\underline{\hspace{0.25cm}}}}
\newcommand\blanktoken{``\underline{\hspace{0.25cm}}''}

\newcommand\tx{\ensuremath{\tilde{x}}}

\newcommand\setsty[1]{\ensuremath{{#1}}}

\newcommand\noised[1]{#1_\gamma}

\newcommand\eqdef{\ensuremath{\stackrel{\rm def}{=}}} 

\ifthenelse{\isundefined{\definition}}{\newtheorem{definition}{Definition}}{}
\ifthenelse{\isundefined{\assumption}}{\newtheorem{assumption}{Assumption}}{}
\ifthenelse{\isundefined{\hypothesis}}{\newtheorem{hypothesis}{Hypothesis}}{}
\ifthenelse{\isundefined{\proposition}}{\newtheorem{proposition}{Proposition}}{}
\ifthenelse{\isundefined{\theorem}}{\newtheorem{theorem}{Theorem}}{}
\ifthenelse{\isundefined{\lemma}}{\newtheorem{lemma}{Lemma}}{}
\ifthenelse{\isundefined{\corollary}}{\newtheorem{corollary}{Corollary}}{}
\ifthenelse{\isundefined{\alg}}{\newtheorem{alg}{Algorithm}}{}
\ifthenelse{\isundefined{\example}}{\newtheorem{example}{Example}}{}
\newcommand{\E}{\ensuremath{\mathbb{E}}} 

\usepackage{caption}
\usepackage{subcaption}
\usepackage{wrapfig}

\usepackage{algorithm}
\usepackage{algpseudocode}
\usepackage{changepage}

\title{Data Noising as Smoothing in Neural Network Language Models}

\author{Ziang Xie, Sida I. Wang, Jiwei Li, Daniel L\'{e}vy, Aiming Nie, Dan Jurafsky, Andrew Y. Ng\\
        Computer Science Department, Stanford University\\
        \texttt{\{zxie,sidaw,danilevy,anie,ang\}@cs.stanford.edu},\\ \texttt{\tt \{jiweil,jurafsky\}@stanford.edu}}

\iclrfinalcopy 

\begin{document}

\maketitle

\begin{abstract}
Data noising is an effective technique
for regularizing neural network models.
While noising is widely adopted
in application domains such as vision and speech,
commonly used noising primitives have not been developed for
discrete sequence-level settings such as language modeling.
In this paper, we derive a connection between input noising
in neural network
language models and smoothing in $n$-gram models.
Using this connection, we draw upon ideas from smoothing
to develop effective noising schemes.
We demonstrate performance gains when applying the proposed schemes to language modeling
and machine translation.
Finally, we provide empirical analysis
validating the relationship between noising and smoothing.
\end{abstract}

\section{Introduction}

Language models are a crucial component in many domains, such
as autocompletion, machine translation, and speech recognition.
A key challenge when performing estimation in language modeling is the
\textit{data sparsity} problem: due to large vocabulary sizes
and the exponential number of possible
contexts, the majority of possible sequences are rarely or never
observed, even for very short subsequences.

In other application domains, data augmentation has been key to improving
the performance of neural network models in the face of insufficient data.
In computer vision, for example,  there exist well-established primitives for synthesizing
additional image data,
such as by rescaling or applying affine distortions to images \citep{LeCun1998,Krizhevsky2012}.
Similarly, in speech recognition adding a background audio track or applying small shifts along the time dimension
has been shown to yield significant gains, especially in noisy settings~\citep{deng2000large,hannun2014deep}.
However, widely-adopted noising primitives have not yet been developed for neural
network language models.

Classic $n$-gram models of language cope with rare and unseen sequences
by using smoothing methods, such as interpolation or absolute discounting~\citep{chen1996empirical}.
Neural network models, however, have no notion of discrete counts, and
instead use distributed representations to combat
the curse of dimensionality \citep{bengio2003neural}.
Despite the effectiveness of distributed representations,
overfitting due to data sparsity remains an issue.
Existing regularization methods, however, are typically applied
to weights or hidden units within the network~\citep{srivastava2014dropout,le2015simple}
instead of directly considering the input data.

In this work, we consider noising primitives as
a form of data augmentation
for recurrent neural network-based language models.
By examining the expected pseudocounts
from applying the noising schemes, we draw connections between
noising and
linear interpolation smoothing. 
Using this connection, we then derive noising schemes that are analogues of more
advanced smoothing methods.
We demonstrate the effectiveness of
these schemes for regularization through experiments on language modeling and machine translation. 
Finally, we validate our theoretical claims by examining
the empirical effects of noising.

\section{Related Work}
Our work can be viewed as a form of data augmentation,
for which to the best of our knowledge
there exists no widely adopted schemes
in language modeling with neural networks.
Classical regularization methods such as $L_2$-regularization are typically applied to the
model parameters, while dropout is applied
to activations which can be along the forward as well as the recurrent
directions~\citep{zaremba2014recurrent,semeniuta2016recurrent,gal2015dropout}.
Others have introduced methods for recurrent neural networks encouraging
the hidden activations to remain
stable in norm, or constraining the recurrent weight matrix to have eigenvalues
close to one~\citep{krueger2015regularizing,arjovsky2015unitary,le2015simple}.
These methods, however, all consider weights and hidden units instead of the
input data, and
are motivated by the vanishing and exploding gradient problem.

Feature noising has been demonstrated to be effective for structured
prediction tasks, and has been interpreted as an explicit
regularizer~\citep{wang2013feature}.
Additionally,~\cite{wager2014altitude}  show that noising can
inject appropriate generative assumptions into discriminative
models to reduce their generalization error, but do not consider sequence models~\citep{wager2016data}.

The technique of randomly zero-masking input word embeddings
for learning sentence representations has been proposed by~\cite{iyyer2015deep},~\cite{kumar2015ask}, and~\cite{dai2015semi},
and adopted by others such as \cite{bowman2015generating}.
However, to the best of our knowledge, no analysis has been provided besides
reasoning that zeroing embeddings may result in a model ensembling effect
similar to that in standard dropout.
This analysis is applicable to classification tasks involving
sum-of-embeddings or bag-of-words models, but does not capture sequence-level effects.
\cite{bengio2015scheduled} also make an empirical observation that the method
of randomly replacing words with fixed probability with a draw from the
uniform distribution
improved performance slightly for an image captioning task;
however, they do not examine why performance improved.

\section{Method}

\subsection{Preliminaries}

We consider language models where given a sequence of indices $X = (x_1, x_2, \cdots, x_T)$,
over the vocabulary $V$,
we model
\begin{equation*}
    p(X) = \prod_{t=1}^T p(x_t | x_{<t})
\end{equation*}

In $n$-gram models, it is not feasible to model the full context
$x_{<t}$ for large $t$ due to the exponential number of possible histories.
Recurrent neural network (RNN) language models can (in theory) model
longer dependencies, since they operate over distributed hidden states
instead of modeling an exponential number of discrete counts
\citep{bengio2003neural,mikolov2012statistical}.

An $L$-layer
recurrent neural network is modeled as
$h^{(l)}_t = f_\theta(h^{(l)}_{t-1}, h^{(l-1)}_t)$,
where $l$ denotes the layer index, $h^{(0)}$ contains the one-hot encoding of $X$,
and in its simplest form
$f_\theta$ applies an affine transformation followed by a nonlinearity.
In this work, we use RNNs with a more complex form of $f_\theta$,
namely long short-term memory (LSTM) units \citep{hochreiter1997long}, which
have been shown to ease training and allow RNNs to capture longer
dependencies.
The output distribution over the vocabulary $V$ at time $t$ is
$p_\theta(x_t|x_{<t}) = \mathrm{softmax}(g_\theta(h^{(L)}_t))$,
where $g: \mathbb{R}^{|h|} \rightarrow \mathbb{R}^{|V|}$
applies an affine transformation.
The RNN is then trained by minimizing over its parameters $\theta$ the
sequence cross-entropy loss
$\ell(\theta) = -\sum_t \log p_\theta(x_t | x_{<t})$,
thus maximizing the likelihood $p_\theta(X)$.

As an extension, we also consider encoder-decoder or
sequence-to-sequence \citep{cho2014learning,sutskever2014sequence} models where given
an input sequence $X$ and output sequence $Y$ of length $T_Y$, we model
\begin{equation*}
    p(Y|X) = \prod_{t=1}^{T_Y} p(y_t | X, y_{<t}).
\end{equation*}
and minimize the loss $\ell(\theta) = -\sum_t \log p_\theta(y_t | X, y_{<t})$.
This setting can also be seen as conditional language modeling, and encompasses
tasks such as machine translation, where $X$ is a source
language sequence and $Y$ a target language sequence, as well as language modeling, where
$Y$ is the given sequence and $X$ is the empty sequence.

\setlength{\abovedisplayskip}{10pt}
\setlength{\belowdisplayskip}{10pt}

\subsection{Smoothing and Noising}

Recall that for a given context length $l$, an $n$-gram model of order $l+1$ is optimal
under the log-likelihood criterion.
Hence in the case where an RNN with finite context achieves near
the lowest possible cross-entropy loss, it behaves like an $n$-gram model.

Like $n$-gram models, RNNs are trained using maximum likelihood, and can easily
overfit~\citep{zaremba2014recurrent}. While generic regularization methods such $L_2$-regularization and dropout are effective,
they do not take advantage of specific properties of sequence modeling.
In order to understand sequence-specific regularization,
it is helpful to examine $n$-gram language models,
whose properties are well-understood.

\paragraph{Smoothing for $n$-gram models}
When modeling $p(x_t|x_{<t})$,
the maximum likelihood estimate $c(x_{<t}, x_t) / c(x_{<t})$
based on empirical counts puts zero probability on unseen
sequences, and thus smoothing is crucial for obtaining good estimates.
In particular, we consider interpolation,
which performs a weighted average between higher and lower order
models. The idea is that when there are not enough observations of
the full sequence, observations of subsequences can help us
obtain better estimates.%
\footnote{For a thorough review of smoothing methods, we defer
to~\cite{chen1996empirical}.}
For example, in a bigram model,
$p_\mathrm{interp}(x_t|x_{t-1}) = \lambda p(x_t|x_{t-1}) +
(1-\lambda)p(x_t)$, 
where $0 \le \lambda \le 1$.

\paragraph{Noising for RNN models}
We would like to apply well-understood smoothing methods such as
interpolation to RNNs, which are also trained using maximum likelihood.
Unfortunately, RNN models have no notion of counts, and we cannot
directly apply one of the usual smoothing methods.
In this section, we consider two
simple noising schemes which we proceed to show correspond to smoothing methods.
Since we can noise the data while training an RNN,
we can then incorporate well-understood generative
assumptions that are known to be helpful in the domain.
First consider the following two noising schemes:

\begin{itemize}[leftmargin=*]
    \item \textbf{unigram noising}\hspace{1em} For each $x_i$ in $x_{<t}$,
with probability $\gamma$ replace $x_i$ with a sample from the unigram
frequency distribution.
    \item \textbf{blank noising}\hspace{1em} For each $x_i$ in $x_{<t}$,
with probability $\gamma$ replace $x_i$ with a placeholder
token \blanktoken.
\end{itemize}

While blank noising can be seen as a way to avoid overfitting
on specific contexts, we will see that both schemes are related to smoothing,
and that unigram noising provides a path
to analogues of more advanced smoothing methods.

\subsection{Noising as Smoothing}
\label{ssec:noisesmooth}

We now consider the maximum likelihood estimate of $n$-gram probabilities
estimated using the pseudocounts of the noised data.
By examining these estimates, we draw a connection between linear interpolation
smoothing and noising.

\paragraph{Unigram noising as interpolation}
To start,  we consider the simplest case of bigram probabilities.
Let $c(x)$ denote the count of a token $x$ in the original data, and let
$\noised{c}(x) \eqdef \E_{\tx}\left[c(\tx)\right]$ be the expected count of $x$
under the unigram noising scheme. We then have
\begin{align*}
    \noised{p}(x_t|x_{t-1}) &= \displaystyle\frac{\noised{c}(x_{t-1}, x_t)}{\noised{c}(x_{t-1})}\\
    &= [(1-\gamma)c(x_{t-1}, x_t) + \gamma\ p(x_{t-1}) c(x_t)] / c(x_{t-1})\\
    &= (1-\gamma)p(x_t | x_{t-1}) + \gamma\ p(x_t),
\end{align*}
where $\noised{c}(x) = c(x)$ since our proposal distribution $q(x)$
is the unigram distribution, and the last line follows since
$c(x_{t-1})/p(x_{t-1}) = c(x_t)/p(x_t)$ is equal to
the total number of tokens in the training set.
Thus we see that the noised data has pseudocounts
corresponding to \emph{interpolation} or a mixture of different order
$n$-gram models with fixed weighting.

More generally, let $\tx_{<t}$ be noised tokens from $\tx$.
We consider the expected prediction under noise
\begin{align*}
    \noised{p}(x_t|x_{<t}) &= \E_{\tx_{<t}}\left[p(x_t|\tx_{<t})\right] \\
    &= \sum_{\setsty{J}} \underbrace{\pi(|\setsty{J}|)}_{p(|J| \text{\ swaps})}  \sum_{x_{\setsty{K}}} \underbrace{p(x_t | x_{\setsty{J}}, x_{\setsty{K}})}_{p(x_t|\text{noised\ context})} \prod_{z \in x_\setsty{K}} \underbrace{p(z)}_{p(\text{drawing\ } z)}
\end{align*}
where the mixture coefficients are
$\pi(|\setsty{J}|) = (1-\gamma)^{|\setsty{J}|} \gamma^{t-1-|\setsty{J}|}$
with $ \sum_{\setsty{J}} \pi( |\setsty{J|}) = 1$.  $\setsty{J} \subseteq \{1,2,\ldots,t-1\}$
denotes the set of indices whose corresponding tokens are left unchanged,
and $\setsty{K}$ the set of indices that were replaced.

\paragraph{Blank noising as interpolation}
Next we consider the blank noising scheme and show that it corresponds to
interpolation as well. This also serves as an alternative explanation for the gains that other related work
have found with the ``word-dropout'' idea~\citep{kumar2015ask,dai2015semi,bowman2015generating}.
As before, we do not noise the token being predicted $x_t$.
Let $\tx_{<t}$ denote the random variable where each of its tokens is
replaced by \blanktoken{} with probability $\gamma$, and let $x_\setsty{J}$ denote the
sequence with indices $\setsty{J}$ unchanged, and the rest replaced by
\blanktoken{}. To make a prediction, we use the expected probability over
different noisings of the context
\begin{equation*}
\noised{p}(x_t|x_{<t}) = \E_{\tx_{<t}}\left[p(x_t|\tx_{<t})\right]
= \sum_{\setsty{J}} \underbrace{\pi(|\setsty{J}|)}_{\hspace{-0.5em} p(|J| \text{\ swaps})} \underbrace{p(x_t | x_{\setsty{J}})}_{p(x_t | \text{noised\ context})},
\end{equation*}
where $\setsty{J} \subseteq \{1,2,\ldots,t-1\}$,
which is also a mixture of the unnoised probabilities over subsequences of the current context.
For example, in the case of trigrams, we have
\begin{align*}
    \noised{p}(x_3|x_1, x_2) =\ &\pi(2)\ p(x_3 |x_1, x_2 ) +  \pi(1)\ p(x_3 |x_1, \blanktok) +  \pi(1)\ p(x_3 |\blanktok, x_2 ) + \pi(0)\ p(x_3 |\blanktok, \blanktok )
\end{align*}
where the mixture coefficient $\pi(i) = (1-\gamma)^i \gamma^{2-i}$\nolinebreak.

\subsection{Borrowing Techniques}

With the connection between noising and smoothing in place, we now consider how we can improve
the two components of the noising scheme by considering:
\begin{enumerate}
    \item Adaptively computing noising probability $\gamma$ to reflect our confidence
       about a particular input subsequence.
    \item Selecting a proposal distribution $q(x)$ that is less naive
        than the unigram distribution by leveraging higher order
        $n$-gram statistics.
\end{enumerate}

\paragraph{Noising Probability}
\label{ssec:absdiscount}
Although it simplifies analysis, there is no reason why we should
choose fixed $\gamma$; we now consider defining an adaptive $\gamma(x_{1:t})$ which
depends on the input sequence. Consider the following bigrams:
\begin{center}
    {``and the''} \hspace{3cm} {``Humpty Dumpty''}
\end{center}

The first bigram is one of the most common in English corpora;
its probability is hence well estimated and should not be
interpolated with lower order distributions.
In expectation, however, using fixed $\gamma_0$ when noising
results in the same lower order interpolation weight $\pi_{\gamma_0}$
for common as well as rare bigrams.
Intuitively, we should define $\gamma(x_{1:t})$ such that commonly
seen bigrams are less likely to be noised.

The second bigram, ``Humpty Dumpty,'' is relatively uncommon, as are
its constituent unigrams.
However, it forms what \cite{brown1992class} term a ``sticky pair'':
the unigram ``Dumpty'' almost always follows the unigram ``Humpty'',
and similarly, ``Humpty'' almost always precedes ``Dumpty''.
For pairs with high mutual information, we 
wish to avoid backing off from the bigram to the unigram distribution.

Let $N_{1+}(x_1, \bullet) \eqdef |\{x_2 : c(x_1, x_2) > 0\}|$
be the number of distinct continutions following $x_1$,
or equivalently the number of bigram types beginning with $x_1$ \citep{chen1996empirical}.
From the above intuitions, we arrive at the \textit{absolute discounting}
noising probability
$$\gamma_\mathrm{AD}(x_1) = \gamma_0 \frac{N_{1+}(x_1, \bullet)}{\sum_{x_2} c(x_1, x_2)}$$
where for $0 \le \gamma_0 \le 1$ we have $0 \le \gamma_\mathrm{AD} \le 1$,
though in practice we can also clip larger noising probabilities to $1$.
Note that this encourages noising of unigrams that precede many possible other
tokens while discouraging noising of common unigrams, since if we ignore
the final token, $\sum_{x_2} c(x_1, x_2) = c(x_1)$.

\paragraph{Proposal Distribution}

\begin{table}[bt]
\begin{center}
\begin{tabular}{c l l l}
\toprule
Noised & $\gamma(x_{1:2})$ & $q(x)$ & Analogue\\
\midrule
$x_1$ & $\gamma_0$ & $q(``\blanktok")=1$ & interpolation \\
$x_1$ & $\gamma_0$ & unigram & interpolation \\
$x_1$ & $\gamma_0 N_{1+}(x_1, \bullet)/c(x_1)$ & unigram & absolute discounting \\
$x_1, x_2$ & $\gamma_0 N_{1+}(x_1, \bullet)/c(x_1)$ & $q(x) \propto N_{1+}(\bullet, x)$ & Kneser-Ney \\
\bottomrule
\end{tabular}
\end{center}
\caption{\textbf{Noising schemes} Example noising schemes and their bigram smoothing analogues.
Here we consider the bigram probability $p(x_1, x_2) = p(x_2|x_1)p(x_1)$.
Notation: $\gamma(x_{1:t})$ denotes the noising probability for a given input sequence $x_{1:t}$,
$q(x)$ denotes the proposal distribution, and $N_{1+}(x, \bullet)$ denotes
the number of distinct bigrams in the training set where $x$ is the first unigram. In all but the last case
we only noise the context $x_1$ and not the target prediction $x_2$.}
\label{tab:noiseschemes}
\end{table}

While choosing the unigram distribution as the proposal
distribution $q(x)$ preserves unigram frequencies, by borrowing from
the smoothing literature we find another distribution performs better.
We again begin with two motivating examples:
\begin{center}
    ``San Francisco''\hspace{3cm}``New York''
\end{center}
Both bigrams appear frequently in text corpora.
As a direct consequence, the unigrams ``Francisco'' and ``York'' also appear frequently.
However, since ``Francisco'' and ``York'' typically follow
``San'' and ``New'', respectively, they should not have high probability
in the proposal distribution as they might if we use unigram frequencies~\citep{chen1996empirical}.
Instead, it would be better to increase
the proposal probability of unigrams with diverse histories,
or more precisely unigrams that complete a large number of bigram types.
Thus instead of drawing from the unigram distribution,
we consider drawing from
$$q(x) \propto N_{1+}(\bullet, x)$$
Note that we now noise the prediction $x_t$ in addition to the context $x_{1:t-1}$.
Combining this new proposal distribution with the discounted $\gamma_\mathrm{AD}(x_1)$
from the previous section, we obtain the noising
analogue of Kneser-Ney smoothing.

Table~\ref{tab:noiseschemes} summarizes the discussed noising schemes.

\subsection{Training and Testing}
During training, noising is performed per batch and is done online such that
each epoch of training sees a different noised version of the training data.
At test time, to match the training objective we should sample multiple
corrupted versions of the test data, then average the predictions~\citep{srivastava2014dropout}.
In practice, however, we find that simply using the maximum
likelihood (uncorrupted) input sequence works well; evaluation runtime remains
unchanged.

\subsection{Extensions}
The schemes described are for the language model setting.
To extend them to the sequence-to-sequence or encoder-decoder setting, we noise both
$x_{<t}$ as well as $y_{<t}$.
While in the decoder we have $y_{<t}$ and $y_t$ as analogues to language model
context and target prediction, it is unclear whether noising $x_{<t}$ should be beneficial.
Empirically, however, we find this to be the case (Table~\ref{tab:mt}).

\section{Experiments}

\subsection{Language Modeling}
\label{ssec:lm}

\paragraph{Penn Treebank}

\begin{table}[bt]
\begin{center}
\begin{tabular}{l r r r }
\toprule
Noising scheme & & Validation & Test \\
\midrule
\multicolumn{4}{c}{Medium models (512 hidden size)}\\
\midrule
\multicolumn{2}{l}{none (dropout only)}
  & 84.3 & 80.4 \\
\multicolumn{2}{l}{blank}
  & 82.7 & 78.8 \\
\multicolumn{2}{l}{unigram}
  & 83.1 & 80.1 \\
\multicolumn{2}{l}{bigram Kneser-Ney}
 & \textbf{79.9} & \textbf{76.9} \\
\midrule
\multicolumn{4}{c}{Large models (1500 hidden size)}\\
\midrule
\multicolumn{2}{l}{none (dropout only)}
  & 81.6 & 77.5 \\
\multicolumn{2}{l}{blank}
  & 79.4 & 75.5 \\
\multicolumn{2}{l}{unigram}
  & 79.4 & 76.1 \\
\multicolumn{2}{l}{bigram Kneser-Ney}
  & \textbf{76.2} & \textbf{73.4} \\
\midrule
\multicolumn{2}{l}{\cite{zaremba2014recurrent}} & 82.2 & 78.4\\
\multicolumn{2}{l}{\cite{gal2015dropout} variational dropout (tied weights)} & 77.3  & 75.0\\
\multicolumn{2}{l}{\cite{gal2015dropout} (untied weights, Monte Carlo)} & {---\hspace{0.15cm}}  & \textbf{73.4}\\
\bottomrule
\end{tabular}
\end{center}
\caption{Single-model perplexity on Penn Treebank with different noising schemes. We also compare
    to the variational method of \cite{gal2015dropout}, who also train LSTM
    models with the same hidden dimension. Note that performing Monte Carlo dropout at test time
    is significantly more expensive than our approach, where test time is unchanged.}
\label{tab:ptb}
\end{table}

\begin{table}[bt]
\begin{center}
\begin{tabular}{l r r }
\toprule
Noising scheme & Validation & Test \\
\midrule
none  & 94.3 & 123.6 \\
blank & 85.0 & 110.7 \\
unigram & 85.2 & 111.3 \\
bigram Kneser-Ney & 84.5 & 110.6 \\
\bottomrule
\end{tabular}
\end{center}
\caption{Perplexity on Text8 with different noising schemes.}
\label{tab:text8}
\end{table}

\begin{figure}[t]
  \centering
  \begin{subfigure}[b]{0.4\textwidth}
      \includegraphics[width=\textwidth]{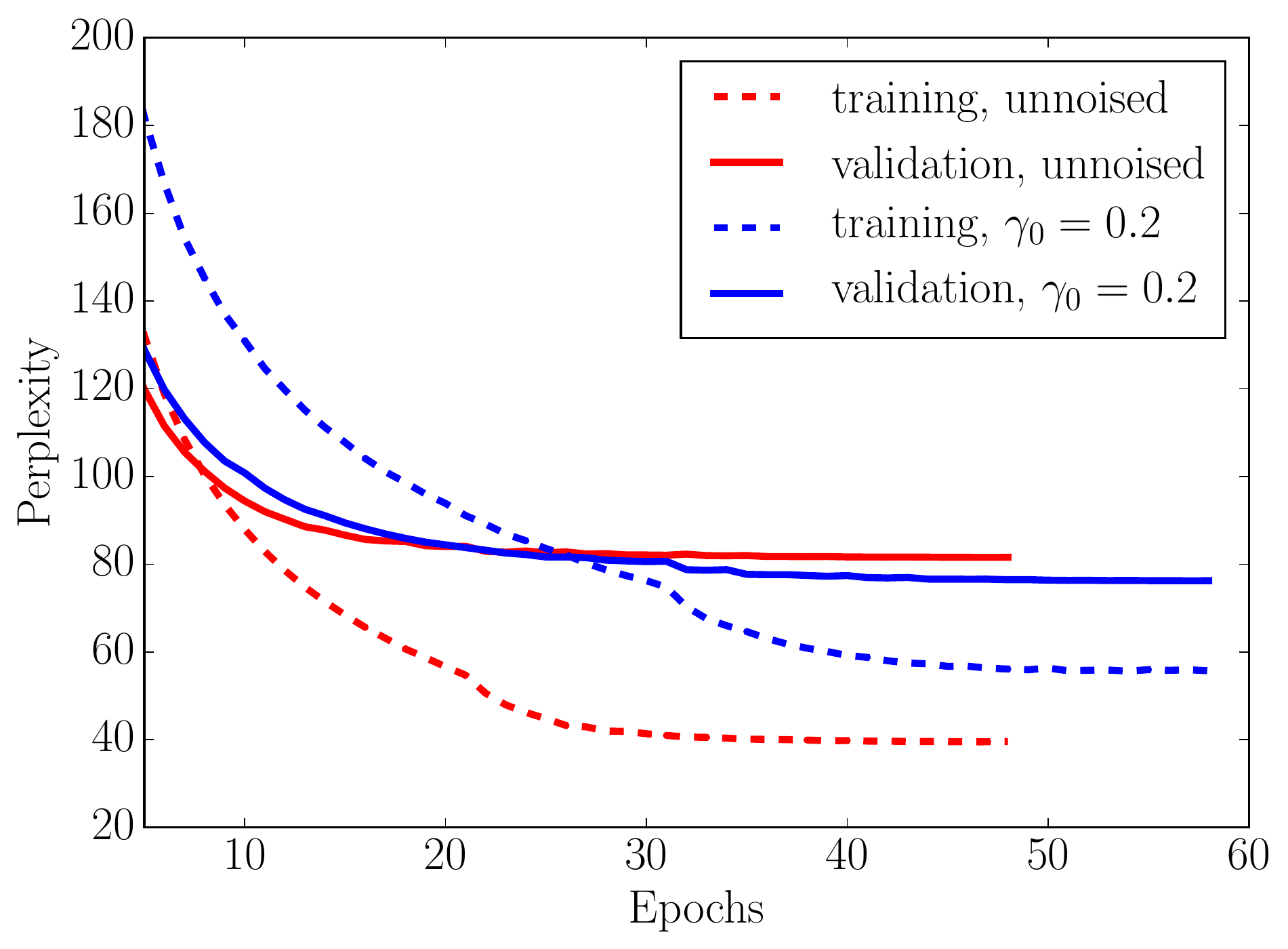}
      \caption{Penn Treebank corpus.}
      \label{fig:ptbcurve}
  \end{subfigure}
  \hspace{1em}
  \begin{subfigure}[b]{0.4\textwidth}
      \includegraphics[width=\textwidth]{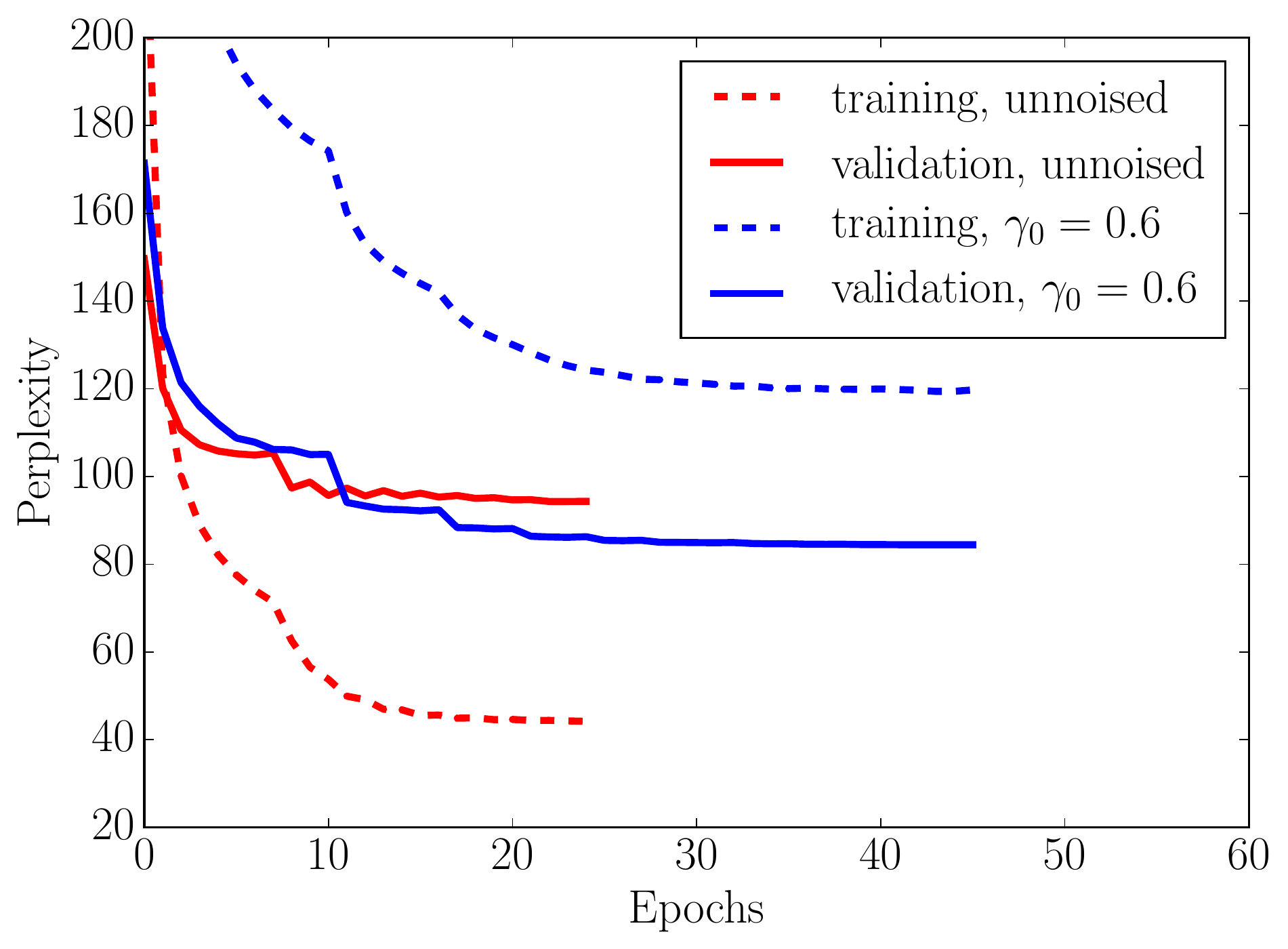}
      \caption{Text8 corpus.}
      \label{fig:text8curve}
  \end{subfigure}
  \caption{Example training and validation curves for an unnoised model and model regularized using the bigram Kneser-Ney noising scheme.}
  \label{fig:trainingcurve}
\end{figure}

We train networks for word-level language modeling on the Penn Treebank dataset,
using the standard preprocessed splits with a 10K size vocabulary~\citep{mikolov2012statistical}.
The PTB dataset contains 929k training tokens, 73k validation tokens, and 82k test tokens.
Following \cite{zaremba2014recurrent},
we use minibatches of size 20 and unroll for 35
time steps when performing backpropagation through time.
All models have two hidden layers and use LSTM units.
Weights are initialized uniformly in the range $[-0.1, 0.1]$.
We consider models with hidden sizes of $512$ and $1500$.

We train using stochastic gradient descent with an initial learning
rate of 1.0, clipping the gradient if its norm exceeds 5.0.
When the validation cross entropy does not decrease after a training epoch,
we halve the learning rate.
We anneal the learning rate 8 times before stopping training, and pick
the model with the lowest perplexity on the validation set.

For regularization, we apply feed-forward dropout~\citep{pham2014dropout}
in combination with our noising schemes.
We report results in Table~\ref{tab:ptb} for the best setting of the dropout rate (which
we find to match the settings reported in~\cite{zaremba2014recurrent})
as well as the best setting of noising probability $\gamma_0$ on the validation set.\footnote{Code will be made available at: \url{http://deeplearning.stanford.edu/noising}}
Figure~\ref{fig:trainingcurve} shows the training and validation perplexity
curves for a noised versus an unnoised run.

Our large models match the state-of-the-art regularization method for single model performance on this task.
In particular, we find that picking $\gamma_\mathrm{AD}(x_1)$ and $q(x)$ corresponding
to Kneser-Ney smoothing yields significant gains in validation perplexity,
both for the medium and large size models.
Recent work ~\citep{merity2016pointer,zilly2016recurrent}
has also achieved impressive results on this task
by proposing different architectures
which are orthogonal to our data augmentation schemes.

\paragraph{Text8}

In order to determine whether noising remains effective with a larger dataset, we
perform experiments on the Text8 corpus\footnote{\url{http://mattmahoney.net/dc/text8.zip}}.
The first 90M characters are used for training,
the next 5M for validation, and the final 5M for testing, resulting in
15.3M training tokens, 848K validation tokens, and 855K test tokens.
We preprocess the data by mapping all words which appear 10 or fewer times
to the unknown token, resulting in a 42K size vocabulary. Other parameter
settings are the same as described in the Penn Treebank experiments, besides
that only models with hidden size 512 are considered, and
noising is not combined with feed-forward dropout. Results are given in
Table~\ref{tab:text8}.

\subsection{Machine Translation}

\begin{table}[bt]
\begin{center}
\begin{tabular}{l l l}
\toprule
Scheme  & Perplexity & BLEU \\
\midrule
dropout, no noising  & 8.84   & 24.6\\
blank noising   & 8.28   & 25.3 ($+0.7$)\\
unigram noising & 8.15 & 25.5 ($+0.9$) \\
bigram Kneser-Ney  & \textbf{7.92} & \textbf{26.0 ($+1.4$)} \\
\midrule
\hfill source only  & 8.74 & 24.8 ($+0.2$) \\
\hfill target only  & 8.14 & 25.6 ($+1.0$) \\
\bottomrule
\end{tabular}
\end{center}
\caption{ Perplexities and BLEU scores for machine translation task. Results
for bigram KN noising on only the source sequence and only the target sequence are given as well.}
\label{tab:mt}
\end{table}

For our machine translation experiments we consider
the English-German machine translation track of IWSLT 2015\footnote{\url{http://workshop2015.iwslt.org/}}.
The IWSLT 2015 corpus consists of sentence-aligned subtitles of TED and TEDx talks.
The training set contains roughly 190K sentence pairs with 5.4M tokens.
Following \cite{luong2015stanford}, we
 use TED tst2012 as a validation set and report
BLEU score results~\citep{papineni2002bleu} on tst2014.
We limit the vocabulary
to the top 50K most frequent words for each language.

We train a two-layer LSTM encoder-decoder network \citep{sutskever2014sequence,cho2014learning}
with $512$ hidden units in each layer.
The decoder uses an attention mechanism~\citep{bahdanau2014neural} with the dot alignment function~\citep{luong2015effective}.
The initial learning rate is 1.0 and
we start halving the learning rate when the relative difference in perplexity on the validation set
between two consecutive epochs is less than $1\%$.
We follow  training protocols as described in \cite{sutskever2014sequence}:
(a) LSTM parameters and word embeddings
are initialized from a uniform distribution
between $[-0.1,0.1]$, (b) inputs
are reversed, (c) batch size is set to 128, (d) gradient clipping is performed
when the norm exceeds a threshold of 5.
We set hidden unit dropout rate to 0.2 across all settings as suggested in \cite{luong2015effective}.
We compare unigram, blank, and bigram Kneser-Ney noising.
Noising rate $\gamma$ is selected on the validation set.

Results are shown in Table~\ref{tab:mt}.
We observe performance gains for both blank noising and unigram noising, giving roughly $+0.7$ BLEU score on the test set. The proposed bigram Kneser-Ney noising scheme gives an additional performance boost of $+0.5$-$0.7$ on top of the blank noising and unigram noising models, yielding a total gain of $+1.4$ BLEU.

\section{Discussion}
\label{sec:analysis}

\begin{figure}[t]
  \centering
  \begin{minipage}{0.48\textwidth}
  \includegraphics[scale=0.75]{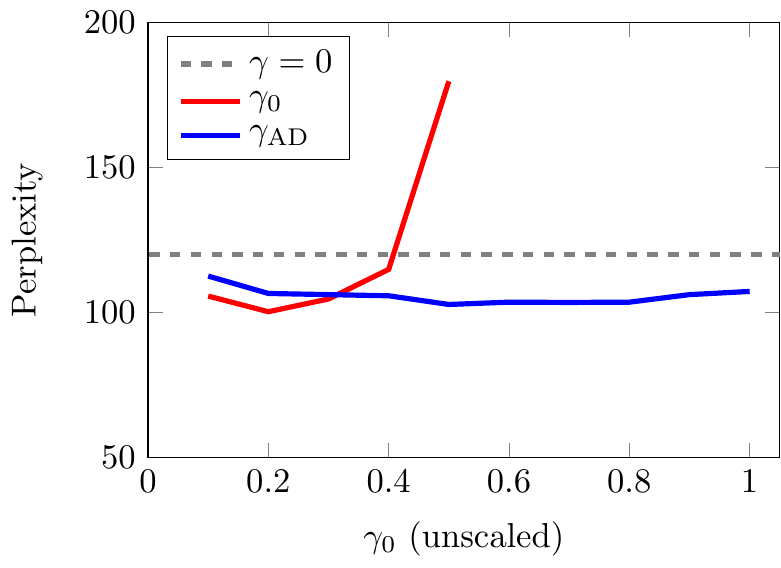}
  \caption{Perplexity with noising on Penn Treebank while varying the value of $\gamma_0$. Using discounting to scale
      $\gamma_0$ (yielding $\gamma_\mathrm{AD}$) maintains gains
      for a range of values of noising probability, which is not true for the unscaled case.}
  \label{fig:discounting}
  \end{minipage}\hspace{0.04\textwidth}%
  \begin{minipage}{0.48\textwidth}
  \vspace{-0.15cm}
  \includegraphics[scale=0.75]{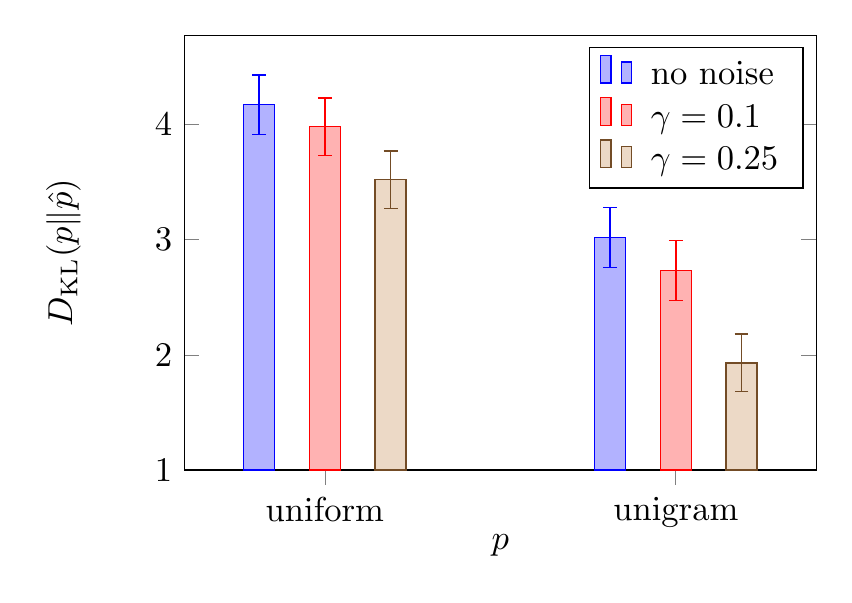}
  \vspace{-0.15cm}
  \caption{Mean KL-divergence over
    validation set between softmax distributions of noised and unnoised models
    and lower order distributions.
    Noised model distributions are closer to the uniform and
    unigram frequency distributions.}
  \label{fig:kls}
  \end{minipage}
\end{figure}

\subsection{Scaling $\gamma$ via Discounting}

We now examine whether discounting has the desired effect of noising
subsequences according to their uncertainty.
If we consider the discounting
$$\gamma_\mathrm{AD}(x_1) =\displaystyle\gamma_0 \frac{N_{1+}(x_1, \bullet)}{c(x_1)}$$
we observe that the denominator $c(x_1)$ can dominate
than the numerator $N_{1+}(x_1, \bullet)$.
Common tokens are often noised infrequently when discounting is used to rescale the
noising probability, while rare tokens are noised comparatively much more frequently,
where in the extreme case when a token appears exactly once, we have $\gamma_\mathrm{AD} = \gamma_0$.
Due to word frequencies following a Zipfian power law distribution, however,
common tokens constitute the majority of most texts, and thus discounting leads
to significantly less noising.

We compare the performance of models trained with a fixed $\gamma_0$ versus
a $\gamma_0$ rescaled using discounting.
As shown in Figure~\ref{fig:discounting}, bigram discounting
leads to gains in perplexity for a much broader range of $\gamma_0$.
Thus the discounting ratio seems to effectively capture the ``right'' tokens to noise.

\subsection{Noised versus Unnoised Models}
\label{ssec:modelcompare}

\paragraph{Smoothed distributions}

In order to validate
that data noising for RNN models has a similar effect to that of smoothing
counts in $n$-gram models, we consider three models trained
with unigram noising as described
in Section~\ref{ssec:lm} on
the Penn Treebank corpus with $\gamma=0$ (no noising),
$\gamma=0.1$, and $\gamma=0.25$.
Using the trained models, we measure the Kullback-Leibler divergence
$D_\mathrm{KL}(p\|q) = \sum_i p_i \log(p_i/q_i)$
over the validation set between
the predicted softmax distributions, $\hat{p}$,
and the uniform distribution
as well as the unigram frequency distribution.
We then take the mean KL divergence over all tokens in the validation set.

Recall that in interpolation smoothing, a weighted combination of
higher and lower order $n$-gram models is used.
As seen in Figure~\ref{fig:kls}, the softmax distributions
of noised models are significantly
closer to the lower order frequency distributions than unnoised models,
in particular in the case of the unigram distribution,
thus validating our analysis in Section~\ref{ssec:noisesmooth}.

\begin{table}
\begin{center}
\begin{tabular}{l r r r}
\toprule
Noising & Bigrams & Trigrams \\
\midrule
none (dropout only) & 2881 & 381\\
blank noising  & 2760 & 372\\
unigram noising & 2612 & 365\\
\bottomrule
\end{tabular}
\end{center}
\caption{Perplexity of last unigram for unseen bigrams and trigrams in
    Penn Treebank validation set. We compare noised and unnoised models
    with noising probabilities chosen such that models have
    near-identical perplexity on full validation set.}
\label{tab:unseen}
\end{table}

\paragraph{Unseen $n$-grams}
Smoothing is most beneficial for increasing the probability of
unobserved sequences. To measure whether noising has a similar
effect, we consider bigrams and trigrams in the
validation set that do not appear in the training set.
For these unseen bigrams (15062 occurrences) and trigrams (43051 occurrences),
we measure the perplexity for noised and unnoised models with near-identical
perplexity on the full set.
As expected, noising yields lower perplexity for these unseen instances.

\section{Conclusion}

In this work, we show that data noising is
effective for regularizing neural network-based sequence models.
By deriving a correspondence between noising
and smoothing, we are able to adapt advanced smoothing methods
for $n$-gram models to the neural network setting,
thereby incorporating well-understood generative assumptions of language.
Possible applications include
exploring noising for improving performance in low resource settings,
or examining how these techniques generalize to sequence modeling in other domains.

\section*{Acknowledgments}

We thank Will Monroe for feedback on a draft of this paper,
Anand Avati for help running experiments, and Jimmy Wu for
computing support.
We also thank the developers of Theano~\citep{2016theano} and Tensorflow~\citep{abadi2016tensorflow}.
Some GPUs used in this work were donated by NVIDIA Corporation.
ZX, SW, and JL were supported by an NDSEG Fellowship, NSERC PGS-D Fellowship,
and Facebook Fellowship, respectively.
This project was funded in part by DARPA MUSE award FA8750-15-C-0242 AFRL/RIKF.


\bibliography{refs}

\begin{thebibliography}{35}
\providecommand{\natexlab}[1]{#1}
\providecommand{\url}[1]{\texttt{#1}}
\expandafter\ifx\csname urlstyle\endcsname\relax
  \providecommand{\doi}[1]{doi: #1}\else
  \providecommand{\doi}{doi: \begingroup \urlstyle{rm}\Url}\fi

\bibitem[Abadi et~al.(2016)Abadi, Agarwal, Barham, Brevdo, Chen, Citro,
  Corrado, Davis, Dean, Devin, et~al.]{abadi2016tensorflow}
Mart{\i}n Abadi, Ashish Agarwal, Paul Barham, Eugene Brevdo, Zhifeng Chen,
  Craig Citro, Greg~S Corrado, Andy Davis, Jeffrey Dean, Matthieu Devin, et~al.
\newblock Tensorflow: Large-scale machine learning on heterogeneous distributed
  systems.
\newblock \emph{arXiv preprint arXiv:1603.04467}, 2016.

\bibitem[Arjovsky et~al.(2015)Arjovsky, Shah, and Bengio]{arjovsky2015unitary}
Martin Arjovsky, Amar Shah, and Yoshua Bengio.
\newblock Unitary evolution recurrent neural networks.
\newblock \emph{arXiv preprint arXiv:1511.06464}, 2015.

\bibitem[Bahdanau et~al.(2014)Bahdanau, Cho, and Bengio]{bahdanau2014neural}
Dzmitry Bahdanau, Kyunghyun Cho, and Yoshua Bengio.
\newblock Neural machine translation by jointly learning to align and
  translate.
\newblock \emph{arXiv preprint arXiv:1409.0473}, 2014.

\bibitem[Bengio et~al.(2015)Bengio, Vinyals, Jaitly, and
  Shazeer]{bengio2015scheduled}
Samy Bengio, Oriol Vinyals, Navdeep Jaitly, and Noam Shazeer.
\newblock Scheduled sampling for sequence prediction with recurrent neural
  networks.
\newblock In \emph{Neural Information Processing Systems (NIPS)}, 2015.

\bibitem[Bengio et~al.(2003)Bengio, Ducharme, Vincent, and
  Jauvin]{bengio2003neural}
Yoshua Bengio, R{\'e}jean Ducharme, Pascal Vincent, and Christian Jauvin.
\newblock A neural probabilistic language model.
\newblock In \emph{Journal Of Machine Learning Research}, 2003.

\bibitem[Bowman et~al.(2015)Bowman, Vilnis, Vinyals, Dai, Jozefowicz, and
  Bengio]{bowman2015generating}
Samuel~R Bowman, Luke Vilnis, Oriol Vinyals, Andrew~M Dai, Rafal Jozefowicz,
  and Samy Bengio.
\newblock Generating sentences from a continuous space.
\newblock \emph{arXiv preprint arXiv:1511.06349}, 2015.

\bibitem[Brown et~al.(1992)Brown, Desouza, Mercer, Pietra, and
  Lai]{brown1992class}
Peter~F Brown, Peter~V Desouza, Robert~L Mercer, Vincent J~Della Pietra, and
  Jenifer~C Lai.
\newblock Class-based n-gram models of natural language.
\newblock \emph{Computational linguistics}, 1992.

\bibitem[Chen \& Goodman(1996)Chen and Goodman]{chen1996empirical}
Stanley~F Chen and Joshua Goodman.
\newblock An empirical study of smoothing techniques for language modeling.
\newblock In \emph{Association for Computational Linguistics (ACL)}, 1996.

\bibitem[Cho et~al.(2014)Cho, Van~Merri{\"e}nboer, Gulcehre, Bahdanau,
  Bougares, Schwenk, and Bengio]{cho2014learning}
Kyunghyun Cho, Bart Van~Merri{\"e}nboer, Caglar Gulcehre, Dzmitry Bahdanau,
  Fethi Bougares, Holger Schwenk, and Yoshua Bengio.
\newblock Learning phrase representations using rnn encoder-decoder for
  statistical machine translation.
\newblock \emph{arXiv preprint arXiv:1406.1078}, 2014.

\bibitem[Dai \& Le(2015)Dai and Le]{dai2015semi}
Andrew~M Dai and Quoc~V Le.
\newblock Semi-supervised sequence learning.
\newblock In \emph{Advances in Neural Information Processing Systems}, pp.\
  3061--3069, 2015.

\bibitem[Deng et~al.(2000)Deng, Acero, Plumpe, and Huang]{deng2000large}
Li~Deng, Alex Acero, Mike Plumpe, and Xuedong Huang.
\newblock Large-vocabulary speech recognition under adverse acoustic
  environments.
\newblock In \emph{ICSLP}, 2000.

\bibitem[Gal(2015)]{gal2015dropout}
Yarin Gal.
\newblock A theoretically grounded application of dropout in recurrent neural
  networks.
\newblock \emph{arXiv:1512.05287}, 2015.

\bibitem[Hannun et~al.(2014)Hannun, Case, Casper, Catanzaro, Diamos,
  et~al.]{hannun2014deep}
Awni Hannun, Carl Case, Jared Casper, Bryan Catanzaro, Greg Diamos, et~al.
\newblock Deep speech: Scaling up end-to-end speech recognition.
\newblock \emph{arXiv preprint arXiv:1412.5567}, 2014.

\bibitem[Hochreiter \& Schmidhuber(1997)Hochreiter and
  Schmidhuber]{hochreiter1997long}
Sepp Hochreiter and J{\"u}rgen Schmidhuber.
\newblock Long short-term memory.
\newblock \emph{Neural computation}, 1997.

\bibitem[Iyyer et~al.(2015)Iyyer, Manjunatha, Boyd-Graber, and
  III]{iyyer2015deep}
Mohit Iyyer, Varun Manjunatha, Jordan Boyd-Graber, and Hal~Daum{\'e} III.
\newblock Deep unordered composition rivals syntactic methods for text
  classification.
\newblock In \emph{Association for Computatonal Linguistics (ACL)}, 2015.

\bibitem[Krizhevsky et~al.(2012)Krizhevsky, Sutskever, and
  Hinton]{Krizhevsky2012}
A.~Krizhevsky, I.~Sutskever, and G.~E. Hinton.
\newblock {ImageNet Classification with Deep Convolutional Neural Networks}.
\newblock In \emph{NIPS}, 2012.

\bibitem[Krueger \& Memisevic(2015)Krueger and
  Memisevic]{krueger2015regularizing}
David Krueger and Roland Memisevic.
\newblock Regularizing rnns by stabilizing activations.
\newblock \emph{arXiv preprint arXiv:1511.08400}, 2015.

\bibitem[Kumar et~al.(2015)Kumar, Irsoy, Su, Bradbury, English, Pierce,
  Ondruska, Gulrajani, and Socher]{kumar2015ask}
Ankit Kumar, Ozan Irsoy, Jonathan Su, James Bradbury, Robert English, Brian
  Pierce, Peter Ondruska, Ishaan Gulrajani, and Richard Socher.
\newblock Ask me anything: Dynamic memory networks for natural language
  processing.
\newblock \emph{arXiv preprint arXiv:1506.07285}, 2015.

\bibitem[Le et~al.(2015)Le, Jaitly, and Hinton]{le2015simple}
Quoc~V Le, Navdeep Jaitly, and Geoffrey~E Hinton.
\newblock A simple way to initialize recurrent networks of rectified linear
  units.
\newblock \emph{arXiv preprint arXiv:1504.00941}, 2015.

\bibitem[LeCun et~al.(1998)LeCun, Bottou, Bengio, and Haffner]{LeCun1998}
Y.~LeCun, L.~Bottou, Y.~Bengio, and P.~Haffner.
\newblock {Gradient-based Learning Applied to Document Recognition}.
\newblock \emph{Proceedings of the IEEE}, 86\penalty0 (11):\penalty0
  2278--2324, 1998.

\bibitem[Luong \& Manning(2015)Luong and Manning]{luong2015stanford}
Minh-Thang Luong and Christopher~D Manning.
\newblock Stanford neural machine translation systems for spoken language
  domains.
\newblock In \emph{Proceedings of the International Workshop on Spoken Language
  Translation}, 2015.

\bibitem[Luong et~al.(2015)Luong, Pham, and Manning]{luong2015effective}
Minh-Thang Luong, Hieu Pham, and Christopher~D. Manning.
\newblock Effective approaches to attention-based neural machine translation.
\newblock In \emph{Empirical Methods in Natural Language Processing (EMNLP)},
  2015.

\bibitem[Merity et~al.(2016)Merity, Xiong, Bradbury, and
  Socher]{merity2016pointer}
Stephen Merity, Caiming Xiong, James Bradbury, and Richard Socher.
\newblock Pointer sentinel mixture models.
\newblock \emph{arXiv preprint arXiv:1609.07843}, 2016.

\bibitem[Mikolov(2012)]{mikolov2012statistical}
Tom{\'a}{\v{s}} Mikolov.
\newblock \emph{Statistical language models based on neural networks}.
\newblock PhD thesis, PhD thesis, Brno University of Technology. 2012.[PDF],
  2012.

\bibitem[Papineni et~al.(2002)Papineni, Roukos, Ward, and
  Zhu]{papineni2002bleu}
Kishore Papineni, Salim Roukos, Todd Ward, and Wei-Jing Zhu.
\newblock Bleu: a method for automatic evaluation of machine translation.
\newblock In \emph{Proceedings of the 40th annual meeting on association for
  computational linguistics}, pp.\  311--318. Association for Computational
  Linguistics, 2002.

\bibitem[Pham et~al.(2014)Pham, Bluche, Kermorvant, and
  Louradour]{pham2014dropout}
Vu~Pham, Th{\'e}odore Bluche, Christopher Kermorvant, and J{\'e}r{\^o}me
  Louradour.
\newblock Dropout improves recurrent neural networks for handwriting
  recognition.
\newblock In \emph{Frontiers in Handwriting Recognition (ICFHR), 2014 14th
  International Conference on}, 2014.

\bibitem[Semeniuta et~al.(2016)Semeniuta, Severyn, and
  Barth]{semeniuta2016recurrent}
Stanislau Semeniuta, Aliaksei Severyn, and Erhardt Barth.
\newblock Recurrent dropout without memory loss.
\newblock \emph{arXiv preprint arXiv:1603.05118}, 2016.

\bibitem[Srivastava et~al.(2014)Srivastava, Hinton, Krizhevsky, Sutskever, and
  Salakhutdinov]{srivastava2014dropout}
Nitish Srivastava, Geoffrey Hinton, Alex Krizhevsky, Ilya Sutskever, and Ruslan
  Salakhutdinov.
\newblock Dropout: A simple way to prevent neural networks from overfitting.
\newblock \emph{The Journal of Machine Learning Research}, 2014.

\bibitem[Sutskever et~al.(2014)Sutskever, Vinyals, and
  Le]{sutskever2014sequence}
Ilya Sutskever, Oriol Vinyals, and Quoc~V Le.
\newblock Sequence to sequence learning with neural networks.
\newblock In \emph{Advances in neural information processing systems}, pp.\
  3104--3112, 2014.

\bibitem[{Theano Development Team}(2016)]{2016theano}
{Theano Development Team}.
\newblock {Theano: A {Python} framework for fast computation of mathematical
  expressions}.
\newblock \emph{arXiv e-prints}, abs/1605.02688, May 2016.
\newblock URL \url{http://arxiv.org/abs/1605.02688}.

\bibitem[Wager et~al.(2014)Wager, Fithian, Wang, and Liang]{wager2014altitude}
S.~Wager, W.~Fithian, S.~I. Wang, and P.~Liang.
\newblock Altitude training: Strong bounds for single-layer dropout.
\newblock In \emph{Advances in Neural Information Processing Systems (NIPS)},
  2014.

\bibitem[Wager et~al.(2016)Wager, Fithian, and Liang]{wager2016data}
Stefan Wager, William Fithian, and Percy Liang.
\newblock Data augmentation via levy processes.
\newblock \emph{arXiv preprint arXiv:1603.06340}, 2016.

\bibitem[Wang et~al.(2013)Wang, Wang, Wager, Liang, and
  Manning]{wang2013feature}
Sida~I Wang, Mengqiu Wang, Stefan Wager, Percy Liang, and Christopher~D
  Manning.
\newblock Feature noising for log-linear structured prediction.
\newblock In \emph{Empirical Methods in Natural Language Processing (EMNLP)},
  2013.

\bibitem[Zaremba et~al.(2014)Zaremba, Sutskever, and
  Vinyals]{zaremba2014recurrent}
Wojciech Zaremba, Ilya Sutskever, and Oriol Vinyals.
\newblock Recurrent neural network regularization.
\newblock \emph{arXiv preprint arXiv:1409.2329}, 2014.

\bibitem[Zilly et~al.(2016)Zilly, Srivastava, Koutn{\'\i}k, and
  Schmidhuber]{zilly2016recurrent}
Julian~Georg Zilly, Rupesh~Kumar Srivastava, Jan Koutn{\'\i}k, and J{\"u}rgen
  Schmidhuber.
\newblock Recurrent highway networks.
\newblock \emph{arXiv preprint arXiv:1607.03474}, 2016.

\end{thebibliography}
\bibliographystyle{iclr2017_conference}

\vfill\pagebreak
\appendix

\section{Sketch of Noising Algorithm}

We provide pseudocode of the noising algorithm corresponding to bigram Kneser-Ney
smoothing for $n$-grams (In the case of sequence-to-sequence tasks,
we estimate the count-based parameters separately for source and target).
To simplify, we assume a batch size of one. The noising
algorithm is applied to each data batch during training. No noising is applied at
test time.
\begin{algorithm}
    \caption{Bigram KN noising (Language modeling setting)}\label{bgkn}

    \begin{adjustwidth}{0.35cm}{0.0cm}
    \textbf{Require} counts $c(x)$, number of distinct continuations $N_{1+}(x, \bullet)$,
    proposal distribution $q(x) \propto N_{1+}(\bullet, x)$\\
    \textbf{Inputs} $X$, $Y$ batch of unnoised data indices, scaling factor $\gamma_0$\\
    \rule{\linewidth}{0.3pt}\vspace{0em}
    \end{adjustwidth}

    \begin{algorithmic}[0]
        \Procedure{NoiseBGKN}{$X,Y$}\Comment{$X=(x_1,\dots,x_t), Y = (x_2,\dots,x_{t+1})$}
        \State $\tilde{X}, \tilde{Y}\gets X, Y$  
        \For{$j=1,\dots,t$}
        \State $\gamma\gets\displaystyle \gamma_0 N_{1+}(x_j, \bullet)/c(x_j)$
        \If{$\sim \mathrm{Bernoulli}(\gamma)$}
        \State $\tilde{x}_j \sim \mathrm{Categorical}(q)$ \Comment{Updates $\tilde{X}$}
          \State $\tilde{y}_j \sim \mathrm{Categorical}(q)$
        \EndIf
    \EndFor\label{euclidendwhile}
    \State \Return $\tilde{X}, \tilde{Y}$ \Comment{Run training iteration with noised batch }
    \EndProcedure
\end{algorithmic}
\end{algorithm}

\end{document}